\documentclass[journal]{IEEEtran}
\IEEEoverridecommandlockouts

\usepackage{cite}
\usepackage{graphics} 
\usepackage{epsfig} 
\usepackage{mathptmx} 
\usepackage{times} 
\usepackage{stfloats}
\usepackage{amsmath} 
\usepackage{amssymb}  
\usepackage{multirow}
\usepackage{booktabs}
\usepackage{fourier-orns}
\usepackage{setspace}
\usepackage{graphicx}
\usepackage{diagbox}
\usepackage[justification=centering]{caption}
\usepackage{float}
\usepackage[section]{placeins}
\usepackage{mathrsfs}
\usepackage[linesnumbered,ruled,vlined]{algorithm2e}

\begin{document}
	
\title{Dense-TNT: Efficient Vehicle Type Classification Neural Network Using Satellite Imagery}

\author{Ruikang~Luo,~\IEEEmembership{Student Member,~IEEE,}
	    Yaofeng~Song,
	    Han~Zhao,~\IEEEmembership{Student Member,~IEEE,}
	    Yicheng~Zhang,~\IEEEmembership{Member,~IEEE,}
	    Yi~Zhang,~\IEEEmembership{Member,~IEEE,}
	    Nanbin~Zhao,~\IEEEmembership{Student Member,~IEEE,}
	    Liping~Huang,~\IEEEmembership{Member,~IEEE,}
        and~Rong~Su,~\IEEEmembership{Senior Member,~IEEE}
        
\thanks{Ruikang Luo is affiliated with Continental-NTU Corporate Lab, Nanyang Technological University, 50 Nanyang Avenue, 639798, Singapore. Email: ruikang001@e.ntu.edu.sg}
\thanks{Yaofeng Song is affiliated with School of Electrical and Electronic Engineering, Nanyang Technological University, 50 Nanyang Avenue, Singapore 639798. Email: song0223@e.ntu.edu.sg}
\thanks{Han Zhao is affiliated with School of Electrical and Electronic Engineering, Nanyang Technological University, 639798, Singapore Email: ZHAO0278@e.ntu.edu.sg}
\thanks{Yicheng Zhang is affiliated with Institute for Infocomm Research (I2R), Agency for Science, Technology and Research (ASTAR), 138632, Singapore Email: zhang$\_$yicheng@i2r.a-star.edu.sg}%
\thanks{Yi Zhang is affiliated with Institute for Infocomm Research (I2R), Agency for Science, Technology and Research (ASTAR), 138632, Singapore Email: Yi Zhang: yzhang120@e.ntu.edu.sg (Zhang$\_$Yi@i2r.a-star.edu.sg)}%
\thanks{Nanbin Zhao is affiliated with School of Electrical and Electronic Engineering, Nanyang Technological University, 50 Nanyang Avenue, Singapore 639798. Email: NANBIN001@e.ntu.edu.sg}%
\thanks{Liping Huang is affiliated with Continental-NTU Corporate Lab, Nanyang Technological University, 50 Nanyang Avenue, 639798, Singapore. Email: liping.huang@ntu.edu.sg}
\thanks{Rong Su is affiliated with Division of Control and Instrumentation, School of Electrical and Electronic Engineering, Nanyang Technological University, 50 Nanyang Avenue, Singapore 639798. Email: rsu@ntu.edu.sg}
}

\markboth{IEEE TRANSACTIONS ON INTELLIGENT TRANSPORTATION SYSTEMS}%
{Shell \MakeLowercase{\textit{et al.}}: Bare Demo of IEEEtran.cls for IEEE Journals}

\maketitle

\begin{abstract}
Accurate vehicle type classification serves a significant role in the intelligent transportation system. It is critical for ruler to understand the road conditions and usually contributive for the traffic light control system to response correspondingly to alleviate traffic congestion. New technologies and comprehensive data sources, such as aerial photos and remote sensing data, provide richer and high-dimensional information. Also, due to the rapid development of deep neural network technology, image based vehicle classification methods can better extract underlying objective features when processing data. Recently, several deep learning models have been proposed to solve the problem. However, traditional pure convolutional based approaches have constraints on global information extraction, and the complex environment, such as bad weather, seriously limits the recognition capability. To improve the vehicle type classification capability under complex environment, this study proposes a novel Densely Connected Convolutional Transformer in Transformer Neural Network (Dense-TNT) framework for the vehicle type classification by stacking Densely Connected Convolutional Network (DenseNet) and Transformer in Transformer (TNT) layers. Three-region vehicle data and four different weather conditions are deployed for recognition capability evaluation. Experimental findings validate the recognition ability of our proposed vehicle classification model with little decay, even under the heavy foggy weather condition.

\end{abstract}

\begin{IEEEkeywords}
deep learning, transformer, remote sensing, vehicle classification.
\end{IEEEkeywords}

\IEEEpeerreviewmaketitle

\section{Introduction}
\IEEEPARstart{V}{ehicle} type classification is one of the most important parts in intelligent traffic system. Vehicle classification results can be contributive towards traffic parameters statistics, regional traffic demand and supply analysis, time-series traffic information prediction\cite{luo2021deep} and transportation facilities usage management\cite{luo2022ast}\cite{song2022sumo}. Combining with appropriate data processing technique, such as missing data imputation and map matching, further traffic management guidance can be provided\cite{huang2022incremental}. Traditional vehicle type classification methods are mainly based on sensors feedback, such as magnetic induction and ultrasonic\cite{amato2019counting}\cite{luo2020traffic}. As the extensive use of UAV surveillance and satellite remote sensing data, image-based solutions towards intelligent traffic system have been rapidly developed. Image processing approaches can be divided into appearance-based methods and deep learning based methods. Appearance-based methods usually generate the 3D parameter model to represent the vehicle for classification. Deep learning based methods apply image recognition algorithms to extract objective features and classify vehicles.

Although remarkable efforts have been made in remote sensing classification tasks, these methods are not ideal when applied to real applications. There are three main limitations on processing remote sensing data. Firstly, high-resolution satellite remote sensing images are expensive and most accessible open source datasets are in low-resolution\cite{wang2018deep}. The poor quality of images constrains the model performance.  Secondly, optical remote sensing images are highly affected by weather conditions\cite{barmpoutis2020review}. Complex weather conditions, such as fog and haze, lead to degraded and blurred images. Thirdly, some modern progressive car design makes distinction boundaries ambiguous. It is essential to determine vehicle types by considering both local and global dependencies, which places higher requirements on the model design.
\begin{figure}[!htb]
	\centering
	\includegraphics[width=0.9\linewidth]{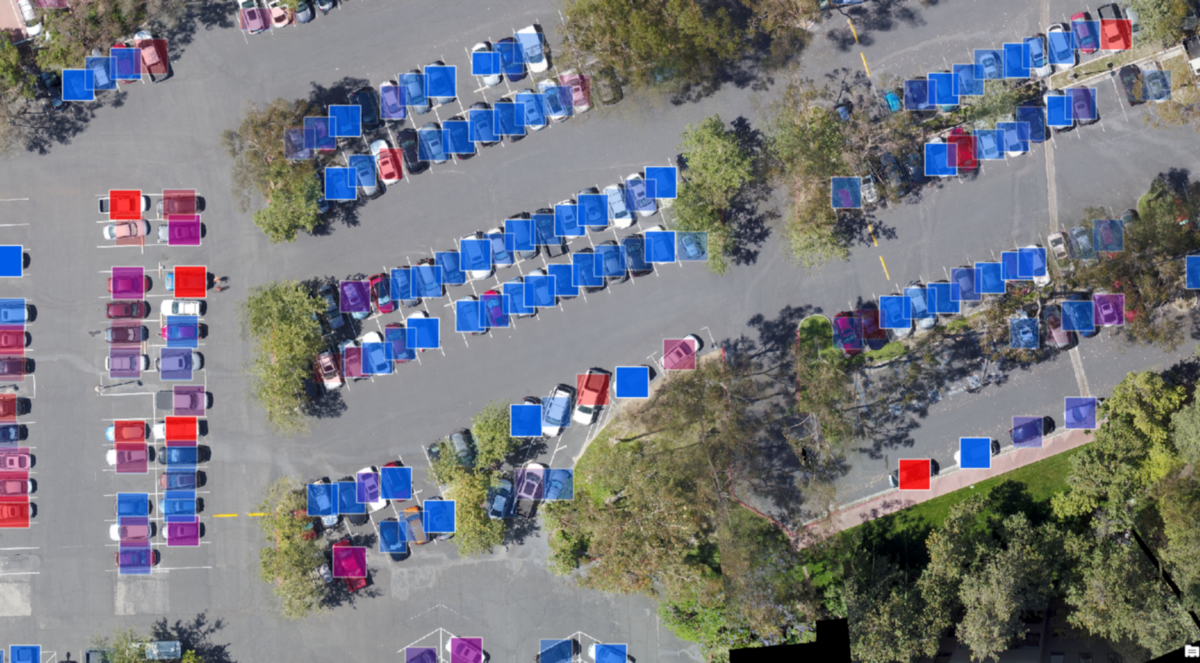}
	\caption{Deep learning algorithms help to capture huge amount of vehicle information in a specific region based on remote sensing data}
	\label{fig00}
\end{figure}

To overcome these issues, existing studies provide solutions in two directions. The first line of work focuses on haze removal or visibility enhancement by utilizing methods, such as Image Super Resolution (ISR)\cite{mahapatra2019image}, to sharpen edges and further improve the resolution. Some methods adopt denoising operations to remove haze and get clearer processed images\cite{wu2019learning}. However, due to the pixels degradation caused by inherent statistical features of fog and haze, haze removal methods may not be effective while processing satellite remote sensing images. The other direction of work tries to build the end-to-end outperformed deep learning algorithm. However, as stated, it is quite a challenge to capture both local and global information under comprehensive conditions.

This study proposes a novel Dense-TNT model for all-weather vehicle classification. It combines a DenseNet layer with a TNT layer, which was recently introduced for objective detection based on the general transformer architecture, to more effectively extract image features. In summary, there are three main contributions as follows:

\begin{itemize}
	\item The novel Dense-TNT model containing the DenseNet layer and TNT layer is proposed to recognize vehicle type. Based on the existing knowledge, the proposed method has the ability to better understand the global pattern of objectives.
	\item The study performs extensive analysis regarding vehicle type classification over remote sensing images collected from three different regions and validate the higher recognition capability than other baselines. 
	\item Apart from three real-world regions data under normal weather condition, appropriate filter is added to simulate the light, medium and heavy haze weather condition. The evaluation results show around 80$\%$ classification accuracy even under heavy-foggy condition and around 5$\%$-10$\%$ accuracy improvement than baseline algorithms. The feasibility has been verified.
\end{itemize}

In the next several sections, the content is organized as follows. In Section Two, recent research on vehicle classification are introduced for comparison. In Section Three, the proposed Dense-TNT framework is described in detail. In Section Four, experiments settings and the vehicle recognition performance is evaluated under various datasets and weather conditions with other baseline models. In Section Five, we give the conclusion and future plan.

\section{Related Work}
Existing vehicle type classification methods can be divided into three categories: appearance-driven methods\cite{ma2005edge}\cite{zhang2006pca}\cite{ji2007vision}\cite{shan2008unsupervised}\cite{jiang2014vehicle}, model-based methods\cite{hsieh2006automatic}\cite{gupte2002detection}\cite{lai2001vehicle} and the deep learning based methods. Appearance-driven methods focus on vehicle appearance features extraction, and try to classify vehicle types by comparing with known vehicle features. In \cite{lowe2004distinctive}, authors propose a method to extract distinctive invariant features and perform robust matching within the known database based on probability indicating. The quantity and quality of known data seriously determine the classification performance and it is difficult for the model to make accurate recognition when the feature of target object is out of database collection, which is common as car designs differ over eras. Model-based methods put efforts on vehicle 3D parameters computation and try to recover the 3D model to make the classification. In \cite{zhang2011three}, a parameterized framework is designed to represent single vehicle with 12 shape parameters and 3 pose parameters. The local gradient-based method is applied to evaluate the goodness of fit between the vehicle projection and known data. However, similar with appearance-driven methods, the appearance and dimension of vehicles can be disturbed and degraded by poor data collection and complex weather condition\cite{pal2018visibility}. Thus, in this study, we mainly discuss the deep learning based methods, which have the capability to capture more information.

For the deep learning based methods, Convolutional Neural Network (CNN) and its variants play a significant role in existing image processing methods\cite{bera2020analysis}. By stacking convolution layer and pooling layer in the classical CNN structure, CNN can automatically learn multi-stage invariant features for the specific objects via trainable kernels\cite{zhang2019recent}. CNN takes the vehicle images as the input and generates each vehicle type probability. However, the pooling operation makes CNN ignore some valuable information without screening the correlation between parts and the entirety carefully\cite{yang2020domain}. Thus, there has also been a lot of interest in combining convolutional layer with attention mechanism for image classification tasks, due to the unbalanced importance distribution over one image\cite{dosovitskiy2020image}. To increase the interpretability of CNN, some research applies semi-supervised manner by feeding unlabeled data in pre-training process and learning output parameters in a supervised way\cite{dong2015vehicle}.

Transformer was proposed in 2017 for Natural Language Processing (NLP) tasks. The principle of attention mechanism leads to the quadratic computational cost by directly applying Transformer on image processing issue, because each pixel needs to attend to every other pixel. Therefore, to adopt Transformer-like structure for image processing, adaptive adjustments are made. In\cite{parmar2018image}, self-attention is applied in local neighborhoods to save operations and replace convolutions\cite{zhao2020exploring}. In another work, Sparse Transformer\cite{child2019generating} uses a scalable filter to global self-attention before processing images. Recently, in the work\cite{yu2022metaformer}, authors replaced the attention mechanism by a token mixer and remained the general Transformer architecture, called MetaFormer. Even combining pooling layer insider token mixer, it is found the model can realize superior performance. Even though CNNs are the fundamental model in vision applications, transformer has a great potential to alternate CNNs.

Vision Transformer (ViT) has been widely used and verified to be efficient in many scenarios, such as object detection, segmentation, pose estimation, image enhancement and video captioning\cite{han2022survey}. Canonical ViT structure divides one image into sequence patches and treat each patch as one element input to do the classification. Due to the inherent characteristic of transformer, ViT is good at long-range relationship extraction, but poor at local features capture since 2D patches are compressed to a 1D vector. Thus, some work tries to improve the local modeling ability\cite{han2021transformer}\cite{liu2021swin}\cite{chen2021regionvit} by introducing extra architecture to model inner correlation patch-by-patch and layer-by-layer\cite{chu2021twins}\cite{lin2021cat}. In \cite{chen2021regionvit}, authors propose a hybrid token generation mechanism to achieve the local and global information from regional tokens and local tokens. In addition to the effort on enhancing local information extraction capability of ViT, some other directions, such as improving self-attention calculation\cite{zhou2021deepvit}, encoding\cite{chu2021conditional}\cite{wu2021rethinking} and normalization strategy\cite{touvron2021going}. In \cite{yu2022metaformer}, authors achieve qualified performance by simplifying the structure even without attention mechanism. 

Due to the fact that low-resolution of data source from satellite imagery and the existence of complex real-world noise, local and global information extraction are both significant for the accurate vehicle classification. Instead of embedding nesting structure within transformer due to the complex computation, in this study, we stack suitable CNN and ViT variants to construct the novel efficient architecture for vehicle classification task. It is expected to achieve satisfying recognition performance, even under complex conditions. If realized, this technology can even be deployed on  nano-satellites, for other earth or celestial bodies recognition tasks\cite{luo2019mission}.

\section{Methodology}
\subsection{Problem Analysis}
In this paper, the main purpose is to build the novel end-to-end vehicle classification model combing selected CNN and ViT variants that takes satellite remote sensing images of various vehicle types from different regions under different weather conditions as inputs, and generates vehicle type classification results after image processing. The principle and detailed architecture of the proposed model is illustrated in this section.

\subsection{Transformer Layer: TNT}
ViT has been successfully applied for wide scenarios and proven to be efficient due to the global long sequence dependencies extraction capability, however, local information aggregation performance is still the gap between ViT and CNNs. Even though some works propose variants that enhance the locality extraction ability, the combination of CNNs and ViT is a more direct method to equip transformer architecture with locality capture.

Similarly, after careful comparison from literature, TNT\cite{han2021transformer} is selected as the variant in our proposed hybrid model. In canonical ViT structure, input images are divided into long sequence patches without local correlation information remained, which it difficult for transformer to catch the relationship simply based on the 2D patch sequence. Comparing with ViT, the main advantages of TNT in terms of this study is introducing another fragment mechanism to create sub-patches within every patch. As the name of TNT, the architecture contains one internal transformer modeling the correlation between sub-patches and one external transformer to propagate information among patches. If the $n$-$length$ patches sequence, $\mathscr{X}^i=\left[X^1, X^2, ..., X^n\right]$, is regarded as visual sentences, each sentence is further divided into $m$ visual words for embedding:
\begin{equation}
X^i \xrightarrow{embedding} Y^i = \left[y^{i,1}, y^{i,2}, ..., y^{i,m}\right]
\end{equation}

For internal transformers, the data flow can be expressed as:
\begin{equation}
{Y'}^{i}_l = Y^i_{l-1} + MSA(LN(Y^i_{l-1}))
\end{equation}
\begin{equation}
{Y}^{i}_l = {Y'}^i_{l} + MLP(LN({Y'}^i_{l}))
\end{equation}
where $l$ is the index of visual words; $MSA$ means Multi-head Self-Attention; $MLP$ means Multi-layer Perceptron and $LN$ represents Layer Normalization. Thus, the overall internal transformation are:
\begin{equation}
\mathscr{Y}_l=\left[Y^1_l, Y^2_l, ..., Y^n_l\right]
\end{equation}

Further, the sequence is transformed from words:
\begin{equation}
Z^i_{l-1} = Z^i_{l-1} + FC(Vec(Y^i_l))
\end{equation}
where $FC$ represents Fully-connected layer. Then, the entire sentence embedding sequence is represented as $\mathscr{Z}_0=\left[Z_{class}, Z^1_0, Z^2_0, ..., Z^n_0\right]$, where $Z_class$ is the class token. Finally, the data flow of the external transformer is formulated as:
\begin{equation}
{\mathscr{Z}'}_l = \mathscr{Z}_{l-1} + MSA(LN(\mathscr{Z}_{l-1}))
\end{equation}
\begin{equation}
\mathscr{Z}_l = {\mathscr{Z}'}_{l} + MLP(LN({\mathscr{Z}'}_{l}))
\end{equation}

In the original paper, authors have shown a better classification performance than other baselines, including ViT. TNT architecture is shown in Fig. 2.

\begin{figure}[!htbp]
	\centering
	\includegraphics[width=1\linewidth]{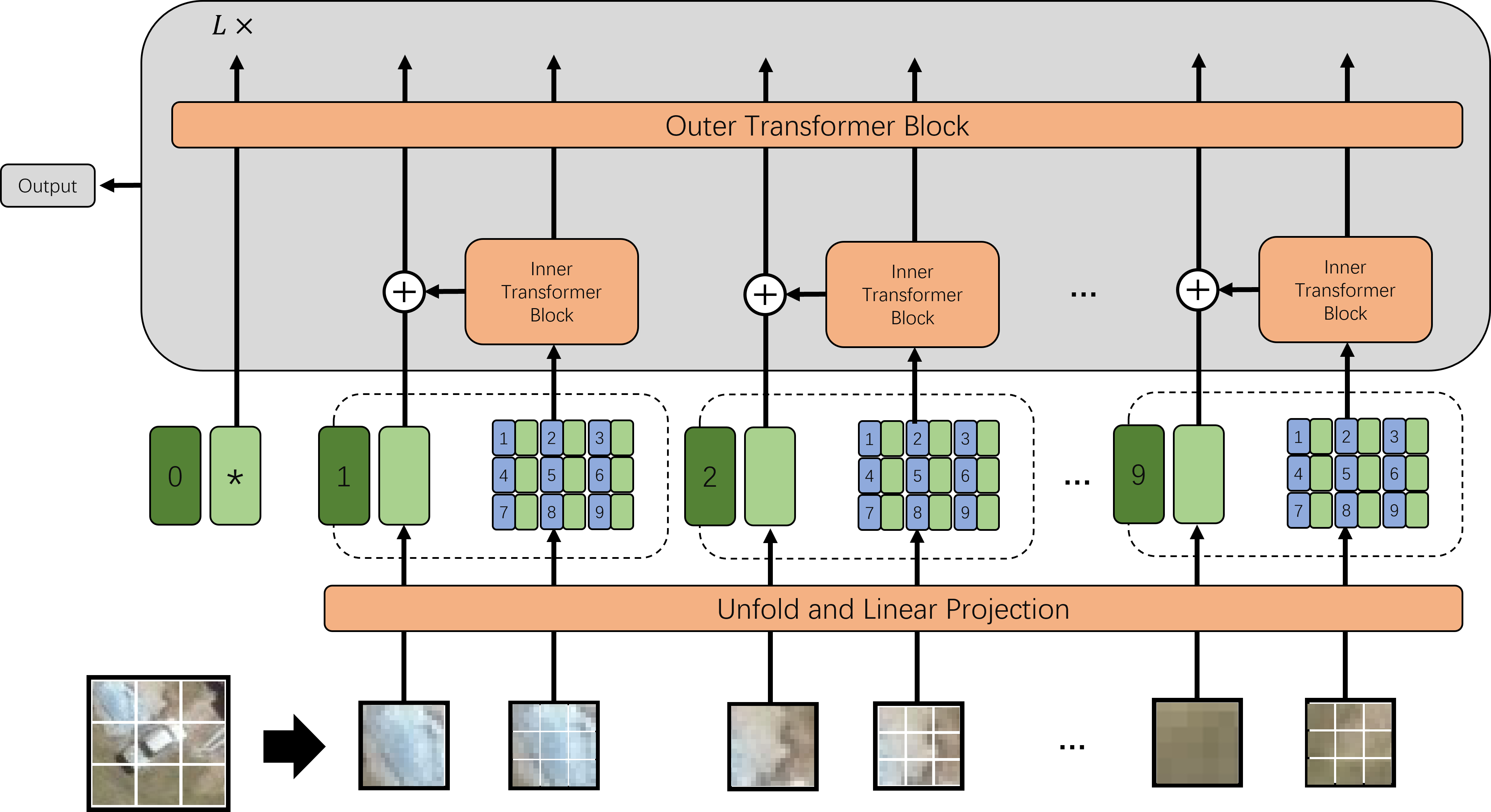}
	\caption{Illustration of TNT model details.}
	\label{fig03}
\end{figure}

\subsection{Convolutional Layer: DenseNet}
For local information extraction part, as introduced in previous sections, CNNs usually show better local fixed information extraction capability due to the existence of kernel structure and convolutional operation\cite{she2018text}. Thus, convolutional layer is remained when designing the efficient image recognition model. In this research, DenseNet is chosen to serve as the locality information extractor. Comparing with other commonly used CNNs models, such as ResNets and GoogLenet, DenseNets connect each convolutional layer with every other layer using the feed-forward network, instead of sequential connections between layers in ResNets and other models. This manner is called dense connectivity in \cite{huang2017densely} and the $i$-$th$ layer is formulated as equation 8:
\begin{equation}
\mathscr{Z}^{*}_i = H_i(\left[\mathscr{Z}_0, \mathscr{Z}_1, ..., \mathscr{Z}_{i-1} \right])
\end{equation}
where $\left[\mathscr{Z}_0, \mathscr{Z}_1, ..., \mathscr{Z}_{i-1} \right]$ is the concatenation result of feature maps from all ahead layers; $H_i(\cdot)$ is the composite function combining batch normalization, rectified linear unit and convolution operations.

In that case, every layer takes features map from all preceding convolutional layers, and information propagation capability is enhanced avoiding serious dependencies loss issue from distant stages. Moreover, gradient vanishing problem is alleviated. In this paper, due to the complex environment of vehicle remote sensing imagery, such as haze weather and shadowy region, the enhanced feature extraction capability is exactly what we need during training. The illustration of DenseNet layout is shown as Fig. 3.
\begin{figure}[!htbp]
	\centering
	\includegraphics[width=0.9\linewidth]{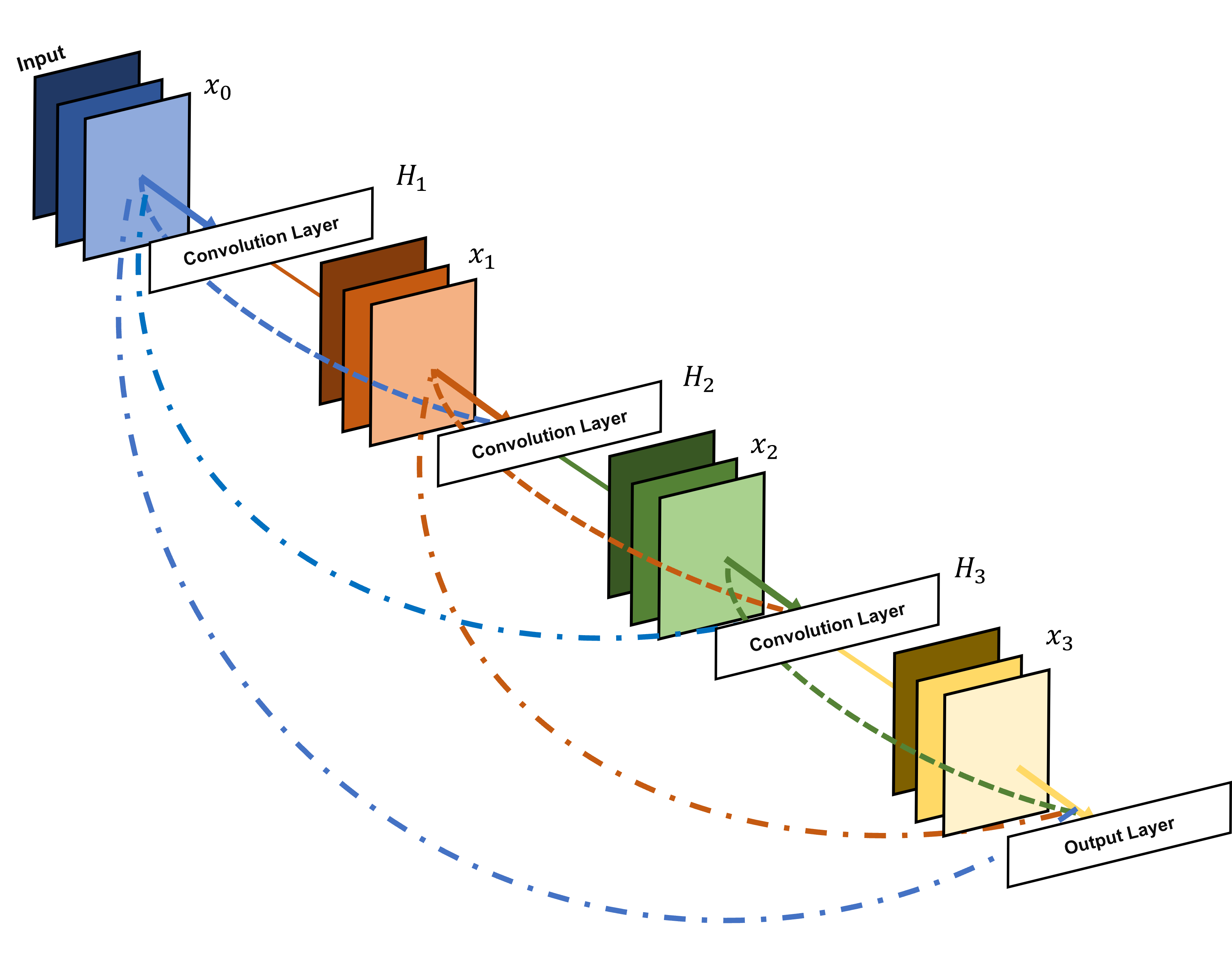}
	\caption{DenseNet model structure: Figure shows a 4-layer dense block. Each layer takes all preceding feature-maps as input. The convolutional layers between two adjacent blocks adjust feature-map sizes.}
	\label{fig02}
\end{figure}

\begin{figure*}[b]
	\centering
	\renewcommand{\dblfloatpagefraction}{0.9}
	\includegraphics[width=1\linewidth]{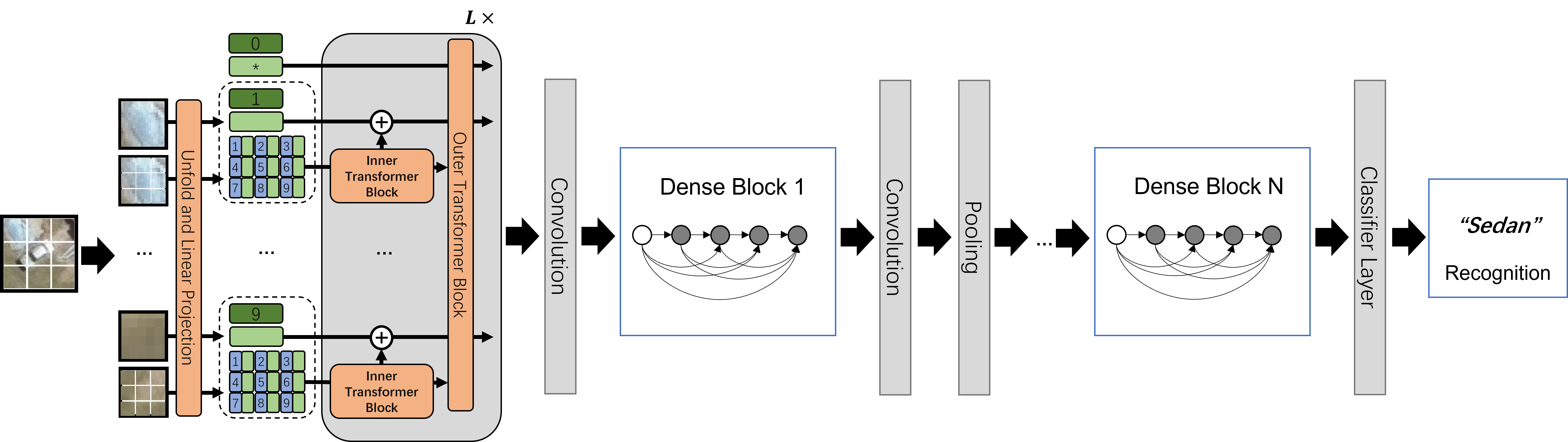}
	\caption{The architecture of Dense-TNT neural network. The network contains TNT and DenseNet stages. The classifier layer serves as the recognition layer to compute the type probability of the input vehicle.}
	\label{fig05}
\end{figure*}

\subsection{Classifier Layer}
To complete the vehicle classification task, the probability of each type is expected to be calculated based on the output feature maps from previous layers. Thus, the softmax classifier layer is added as the final part of our proposed model to take the output feature vector from TNT layer and generate vehicle type probability vector for choice with highest probability. The learnable linear function modeling the relationship can be expressed as:
\begin{equation}
v=W^T\mathscr{Z}^{*}+b
\end{equation}
where $x \in R^{D\times 1}$ is the output feature with $D$ dimension from TNT; $W$ is the parameter to be learned; $v \in R^{C\times 1}$ is the vehicle type variable, and $C$ is the number of vehicle type. To emphasize the vehicle type with the highest probability, softmax is applied to achieve the final normalized output $O=\left[O_1, O_2, ..., O_C\right]^T$:
\begin{equation}
V=\sum_{i=1}^{C}e^{v_i}
\end{equation}
\begin{equation}
O_i=\frac{1}{V}e^{v_i}
\end{equation}

\subsection{Dense-TNT Overview Model}
In summary, the novel Dense-TNT model is designed based on DenseNet and TNT as shown in Fig. 4. It contains two parts: 1) the transformer-based layer, which guarantees baseline reasonable performance 2) the convolutional layer, which captures local fixed features. Beneficial from the kernel and convolutional operation, DenseNet is widely used for image recognition and has deeper locality extraction capability than other CNN variants. Further, TNT is adept in global information capture and has better understanding than canonical ViT. We believe that Dense-TNT can process the information propagated through the hybrid structure by extracting some specific local features and further improve the recognition capability even in the complex environment, such as haze and foggy condition.

\section{Experiments}
To evaluate our Dense-TNT model, we will compare the classification performance of proposed Dense-TNT with several advanced baselines, including PoolFormer and ViT. The main task of classification is to classify sedans and pickups from pictures taken by remote sensors from 3 different areas. Meanwhile, we will also evaluate the classification ability of Dense-TNT when input pictures are affected by fog.

\subsection{Classification in Normal Weather Condition}
\subsubsection{Data Description}
We use Cars Overhead With Context (COWC) (http://gdo-datasci.ucllnl.org/cowc/)\cite{https://doi.org/10.48550/arxiv.1609.04453}, a remote sensing target detection dataset with a resolution of 15cm per pixel and an image size of 64x64, to do the classification. Remote sensing pictures from three different areas, including Toronto Canada, Selwyn New Zealand and Columbus Ohio, are selected. Details of different area datasets are described in Table \uppercase\expandafter{\romannumeral1} and example pictures of sedans and pickups are shown in Table \uppercase\expandafter{\romannumeral2}.

\begin{table}[!htbp]\large
	\renewcommand\arraystretch{1.3}
	\centering
	\resizebox{0.48\textwidth}{!}{%
        \begin{tabular}{cccc}
		\hline
		Locations & Total Number & Number of Sedans & Number of Pickups\\
		\hline
		Columbus Ohio & 7465 & 6917 & 548\\
		Selwyn & 4525 & 3548 & 1067\\
		Toronto & 45994 & 44208 & 1789\\
		\hline
		\end{tabular}%
	    }
    \caption{Details of 3 datasets. The first column refers to 3 different areas where images are taken. The second column refers to the total number of images in the area. The other 2 columns refers to the number of sedans and pickups in the dataset respectively.}	
    \label{table1}
\end{table}

\begin{table}[!htbp]
	\renewcommand\arraystretch{1.3}
	\centering
	\resizebox{0.48\textwidth}{!}{%
		\begin{tabular}{cc}
			\hline
			Sedan & Pickup\\
			\hline
			\begin{minipage}[b]{0.3\columnwidth}
				\centering
				\raisebox{-.5\height}{\includegraphics[width=\linewidth]{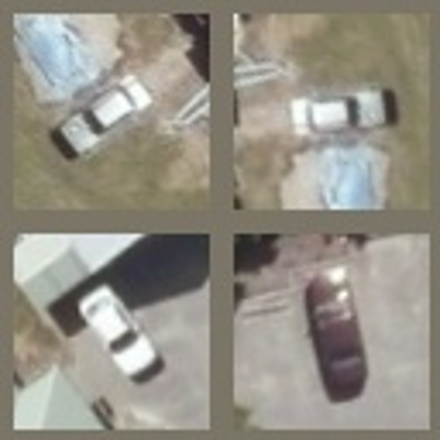}}
			\end{minipage}
			&
			\begin{minipage}[b]{0.3\columnwidth}
				\centering
				\raisebox{-.5\height}{\includegraphics[width=\linewidth]{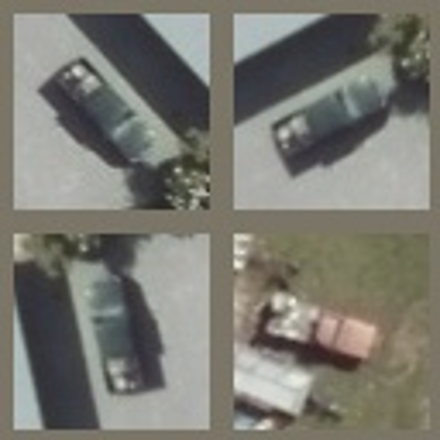}}
			\end{minipage}
			\\
			\hline
		\end{tabular}%
	}
	\caption{Example pictures of sedans and pickups. The first column shows four example pictures of sedans and the second column shows four example pictures of sedans.}	
	\label{table2}
\end{table}

\subsubsection{Baseline Settings}
Baseline models include PoolFormer and ViT. PoolFormer is the specific framework proposed in \cite{yu2022metaformer} and achieves the best recognition performance. ViT is widely applied in image processing problems recently as stated in Section Two.

\subsubsection{Training}
The model is trained with 50 epochs on a RTX3060 GPU with a max learning rate of $lr = 2e^{-3}$. AdamW optimizer is used with weight decay 0.05. The batch size is set to be 0.01 of the size of training data. The size of training data is 0.8 of the size of the whole dataset while the size of test data is 0.2 of the size of the whole dataset. All the training data and the test data are randomly selected from the whole set. Considering the balance of the number of sedans and pickups, we randomly select the same number of sedan and pickup pictures.
\begin{algorithm}{
		\emph {Initialize: input figures preprocessing}\;
		\emph {Initialize: network weights with random values}\;
		\For{$episode=1$ \KwTo $max-episodes$}{
			\For{$\text{batch}=1$ \KwTo $max-\text{batch size}$}{
				\emph {extract input features with TNT}\;
				\emph {classify vehicle types with DenseNets}\;
				\emph {output classification results}\;
				\emph {calculate loss using RMSE}\;
			}
			\emph {perform a gradient descent and network weights update}\;
		}
	}
	\caption{Dense-TNT Model Training}
	\label{Algorithm1}
\end{algorithm}

\subsubsection{Evaluation Criteria}
To evaluate classification outcomes for all models, four criteria are applied to indicate the performance. We first define all classification results in this study as:
\begin{itemize}
	\item True Positive (TP): A sedan is successfully recognized as a sedan.
	\item True Negative (TN): A pickup is successfully recognized as a pickup.
	\item False Positive (FP): A pickup is successfully recognized as a sedan.
	\item False Negative (FN): A sedan is successfully recognized as a pickup.
\end{itemize}

Thus, four criteria are formulated as:
\begin{equation}
Accuracy = \frac{TP+TN}{TP+TN+FP+FN}
\end{equation}
\begin{equation}
Precision = \frac{TP}{TP+FP}
\end{equation}
\begin{equation}
Recall = \frac{TP}{TP+FN}
\end{equation}
\begin{equation}
F1-score = \frac{2 \times Precision \times Recall}{Precision+Recall}
\end{equation}
The higher the evaluation score, the better the recognition performance.

\subsubsection{Experiment Result and Analysis}
We apply Dense-TNT with 2 sizes of parameters, s12 and s24, PoolFormer with 2 sizes of parameters, s12 and s24, and ViT with different numbers of layers, 2 and 12, to do the evaluation. In all the models, 12-layer ViT is equipped with the largest amount of parameters which is about 86M. The parameter numbers of Dense-TNT s24 and PoolFormer s24 are both about 21M. The parameter number of Dense-TNT s12 and PoolFormer s12 are both 12M. Comparing with them, 2-layer ViT has the smallest amount of parameters. Table \uppercase\expandafter{\romannumeral3} shows the results of experiments. Fig. 5 shows the classification results with probabilities after processing under the normal weather condition. 
\begin{figure}[!htb]
	\centering
	\includegraphics[width=0.9\linewidth]{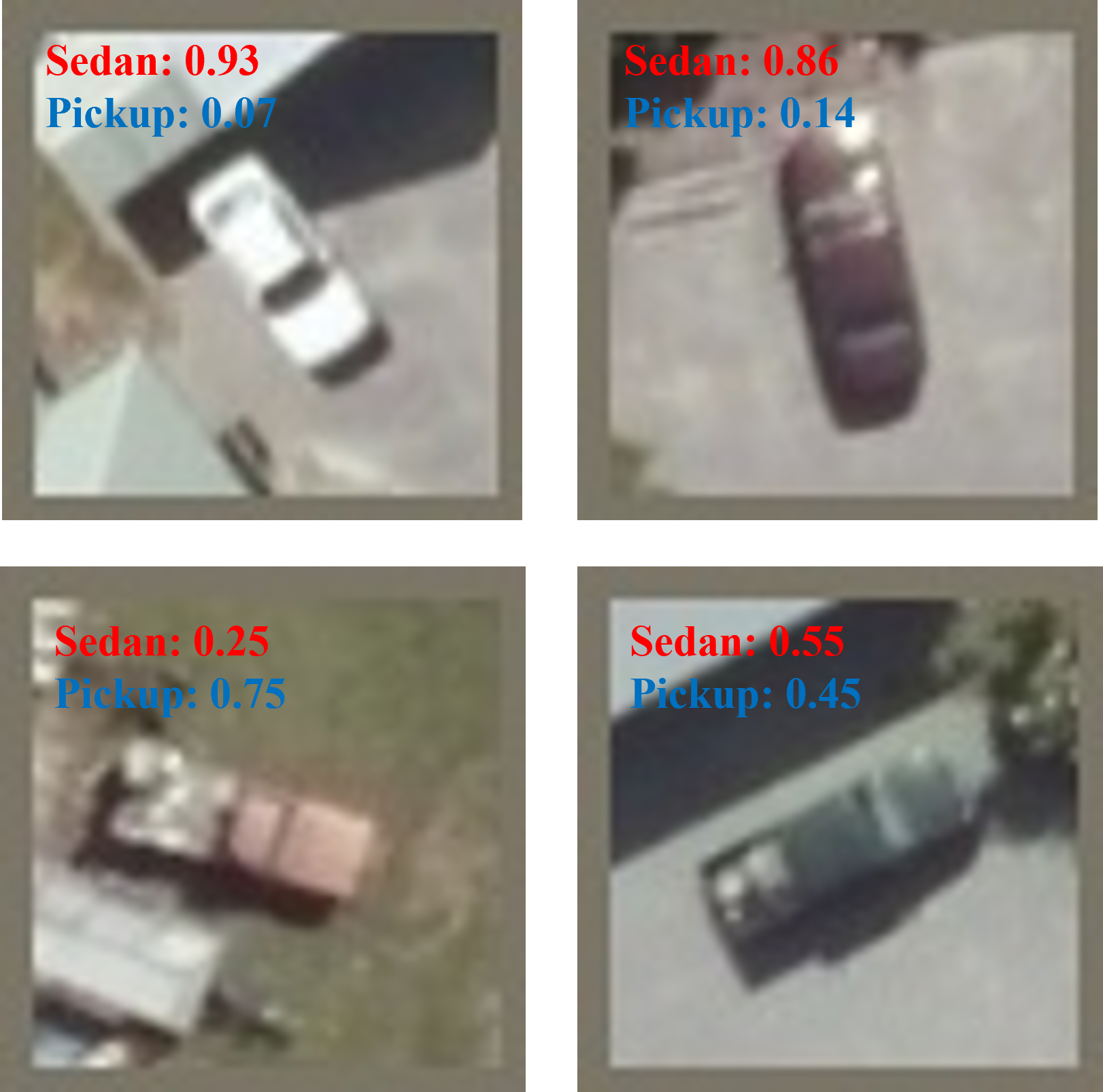}
	\caption{Classification results with corresponding probabilities under the normal weather condition.}
	\label{fig31}
\end{figure}

Through the comparison between ViT l12, Dense-TNT s24 and PoolFormer s24,  Dense-TNT achieves a generally better performance on all the datasets. Even though the computation cost of Dense-TNT s24and PoolFormer s24 are smaller than that of ViT l12, the accuracy of Dense-TNT s24 and PoolFormer s24 are relatively higher. Through the comparison between smaller parameter-size models which includes Dense-TNT s12, PoolFormer s12 and ViT l2, Dense-TNT can also performs better than the other 2 models. With a larger amount of parameters, Dense-TNT s24 has a relatively better performance than Dense-TNT s12.

F1-score is the harmonic mean of precision and recall, which reflects the robustness of model recognition capability. Fig. 6 also shows F1-scores of experiments in histogram and it can be observed that Dense-TNT has prior performance.

\begin{table*}[!htp]\large
	\renewcommand\arraystretch{2}
	\centering
	\resizebox{1\textwidth}{!}{%
	\begin{tabular}{|c|c|c|c|c|c|c|c|c|c|c|c|c|}
		\hline
		\multirow{2}*{\diagbox{\textbf{Models}}{\textbf{Criteria}}} & \multicolumn{4}{c|}{\textbf{Selwyn}} & \multicolumn{4}{c|}{\textbf{Columbus Ohio}} & \multicolumn{4}{c|}{\textbf{Toronto}}\\
		\cline{2-13}
		& Accuracy & Precision & Recall & F1-score & Accuracy & Precision & Recall & F1-score & Accuracy & Precision & Recall & F1-score\\
		\hline
		\textbf{Dense-TNT s24} & \textbf{0.8065} & \textbf{0.8211} & 0.9558 & \textbf{0.8810} & \textbf{0.7685} & \textbf{0.7876} & 0.9516 & 0.8582 & \textbf{0.8009} & 0.8205 & \textbf{0.9365} & \textbf{0.8734} \\
		\hline
		\textbf{Dense-TNT s12} & 0.7971 & 0.8183 & 0.9399 & 0.8722 & 0.7459 & 0.7855 & 0.9109 & 0.8377 & 0.7968 & \textbf{0.8389} & 0.9062 & 0.8706 \\
		\hline
		\textbf{PoolFormer s24} & 0.7819 & 0.7956 & \textbf{0.9559} & 0.8672 & 0.7675 & 0.7835 & \textbf{0.9691} & \textbf{0.8634} & 0.7584 & 0.7871 & 0.9183 & 0.8469 \\
		\hline
		\textbf{PoolFormer s12} & 0.7724 & 0.7977 & 0.9424 & 0.8619 & 0.7507 & 0.7812 & 0.9661 & 0.8431 & 0.7456 & 0.7509 & 0.9254 & 0.8441 \\
		\hline
		\textbf{ViT l12} & 0.7462 & 0.7462 & 0.9256 & 0.8252 & 0.7392 & 0.7421 & 0.9543 & 0.8455 & 0.7300 & 0.7349 & 0.9326 & 0.8401 \\
		\hline
		\textbf{ViT l2} & 0.7624 & 0.7659 & 0.9435 & 0.8623 & 0.7460 & 0.7486 & 0.9339 & 0.8504 & 0.7510 & 0.7559 & 0.9273 & 0.8560\\
		\hline
	\end{tabular}
    }
    \caption{Experiment Results. The table shows the classification accuracy of all the 6 models in the experiments on 3 datasets.}	
    \label{table3}
\end{table*}

\begin{table*}
	\renewcommand\arraystretch{1.3}
	\centering
	\resizebox{0.9\textwidth}{!}{%
		\begin{tabular}{cccc}
			\hline
			Origin & Light & Medium & Heavy\\
			\hline
			\begin{minipage}[b]{0.3\columnwidth}
				\centering
				\raisebox{-.5\height}{\includegraphics[width=\linewidth]{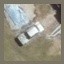}}
			\end{minipage}
			&
			\begin{minipage}[b]{0.3\columnwidth}
				\centering
				\raisebox{-.5\height}{\includegraphics[width=\linewidth]{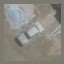}}
			\end{minipage}
			&
			\begin{minipage}[b]{0.3\columnwidth}
				\centering
				\raisebox{-.5\height}{\includegraphics[width=\linewidth]{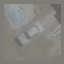}}
			\end{minipage}
			&
			\begin{minipage}[b]{0.3\columnwidth}
				\centering
				\raisebox{-.5\height}{\includegraphics[width=\linewidth]{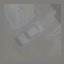}}
			\end{minipage}
			\\
			\hline
		\end{tabular}%
	}
	\caption{Experiment images under different weather conditions. The four columns respectively refers to images taken under normal weather condition, light foggy condition, medium foggy condition and heavy foggy condition.}	
	\label{table4}
\end{table*}
\begin{figure}[!htb]
	\centering
	\includegraphics[width=1\linewidth]{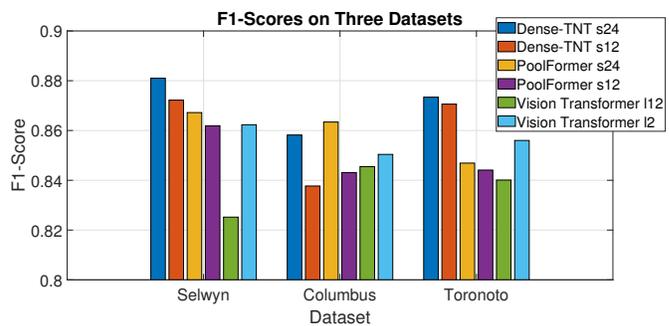}
	\caption{F1-scores of experimental results for three-region datasets under the normal weather condition.}
	\label{fig33}
\end{figure}


\subsection{Classification in Foggy Condition}
\subsubsection{Data Preprocessing}
To achieve vehicle images under different fog conditions, we process the grayscale value of every pixel in the vehicle image. Based on the gray scale value $V_{ij}$ of the pixel on the $i$-$th$ row and $j$-$th$ column in the original image, the new different gray scale values $V_{ij}^{'}$ are achieved by the parameter $\beta$:
\begin{equation}
d = -0.04 \times \sqrt{(i-i_0)^2+(j-j_0)^2}+\sqrt{max(N_R, N_C)}
\end{equation}
\begin{equation}
t_d = e^{-\beta \times d}
\end{equation}
\begin{equation}
V_{ij}^{'} = V_{ij} \times t_d + 0.5 \times (1-t_d)
\end{equation}
where $i_0$ and $j_0$ refer to the center of each row and column respectively, and $N_R$ and $N_C$ refer to the number of pixels in each row and column respectively. In the experiment, parameter $\beta$ is chosen as 0.08, 0.16 and 0.24 to realize 3 types of fog conditions which includes light foggy, medium foggy and heavy foggy. Dataset collected in Selwyn is randomly chosen for the experiments. Table \uppercase\expandafter{\romannumeral4} are example pictures under different levels of weather impact.

\subsubsection{Baseline Settings}
We keep the model size in this experiment the same as the sizes of Dense-TNT models in the 3-area classification in normal weather condition. Models include Dense-TNT s24 and Dense-TNT s12. And other baselines are also kept the same as the previous experiment.
\begin{table*}\large
	\renewcommand\arraystretch{2}
	\centering
	\resizebox{1\textwidth}{!}{%
		\begin{tabular}{|c|c|c|c|c|c|c|c|c|c|c|c|c|}
			\hline
			\multirow{2}*{\diagbox{\textbf{Models}}{\textbf{Criteria}}} & \multicolumn{4}{c|}{\textbf{Light-foggy (fog=0.08)}} & \multicolumn{4}{c|}{\textbf{Medium-foggy (fog=0.16)}} & \multicolumn{4}{c|}{\textbf{Heavy-foggy} (fog=0.24)}\\
			\cline{2-13}
			& Accuracy & Precision & Recall & F1-score & Accuracy & Precision & Recall & F1-score & Accuracy & Precision & Recall & F1-score\\
			\hline
			\textbf{Dense-TNT s24} & \textbf{0.7941} & 0.8240 & 0.9215 & \textbf{0.8682} & \textbf{0.7961} & \textbf{0.7934} & 0.9352 & \textbf{0.8712} & \textbf{0.7692} & 0.7660 & 0.9440 & \textbf{0.8787}\\
			\hline
			\textbf{Dense-TNT s12} & 0.7907 & \textbf{0.8244} & 0.9178 & 0.8671 & 0.7839 & 0.7815 & 0.9510 & 0.8382 & 0.7648 & \textbf{0.7748} & 0.9497 & 0.8715\\
			\hline
			\textbf{PoolFormer s24} & 0.7665 & 0.7912 & 0.9243 & 0.8490 & 0.7590 & 0.7630 & 0.9594 & 0.8608 & 0.7535 & 0.7663 & 0.9543 & 0.8635\\
			\hline
			\textbf{PoolFormer s12} & 0.7631 & 0.7630 & 0.9289 & 0.8641 & 0.7500 & 0.7469 & \textbf{0.9624} & 0.8543 & 0.7371 & 0.7370 & 0.9601 & 0.8469\\
			\hline
			\textbf{ViT l12} & 0.7533 & 0.7539 & \textbf{0.9310} & 0.8569 & 0.7456 & 0.7369 & 0.9449 & 0.8431 & 0.7428 & 0.7400 & 0.9627 & 0.8512\\
			\hline
			\textbf{ViT l2} & 0.7566 & 0.7495 & 0.7297 & 0.8585 & 0.7394 & 0.7402 & 0.9573 & 0.8482 & 0.7383 & 0.7369 & \textbf{0.9659} & 0.8471\\
			\hline
		\end{tabular}
	}
	\caption{Results of experiments with data affected by fog. The first column of the table refers to different models. The other columns shows the accuracy of those 6 models in normal weather condition, light-foggy weather, medium-foggy weather and heavy-foggy weather respectively.}	
	\label{table5}
\end{table*}

\subsubsection{Training}
The training method is also kept unchanged while the input data for model training are replaced with fog affected pictures. The total number of training epoch is 50 on a RTX 3060 GPU. The maximum learning rate is $lr = 2e^{-3}$  and the optimizer is AdamW. The batch size is set to be 0.01 of the size of training data.

\subsubsection{Results}
\begin{figure}[!htbp]
	\centering
	\includegraphics[width=0.7\linewidth]{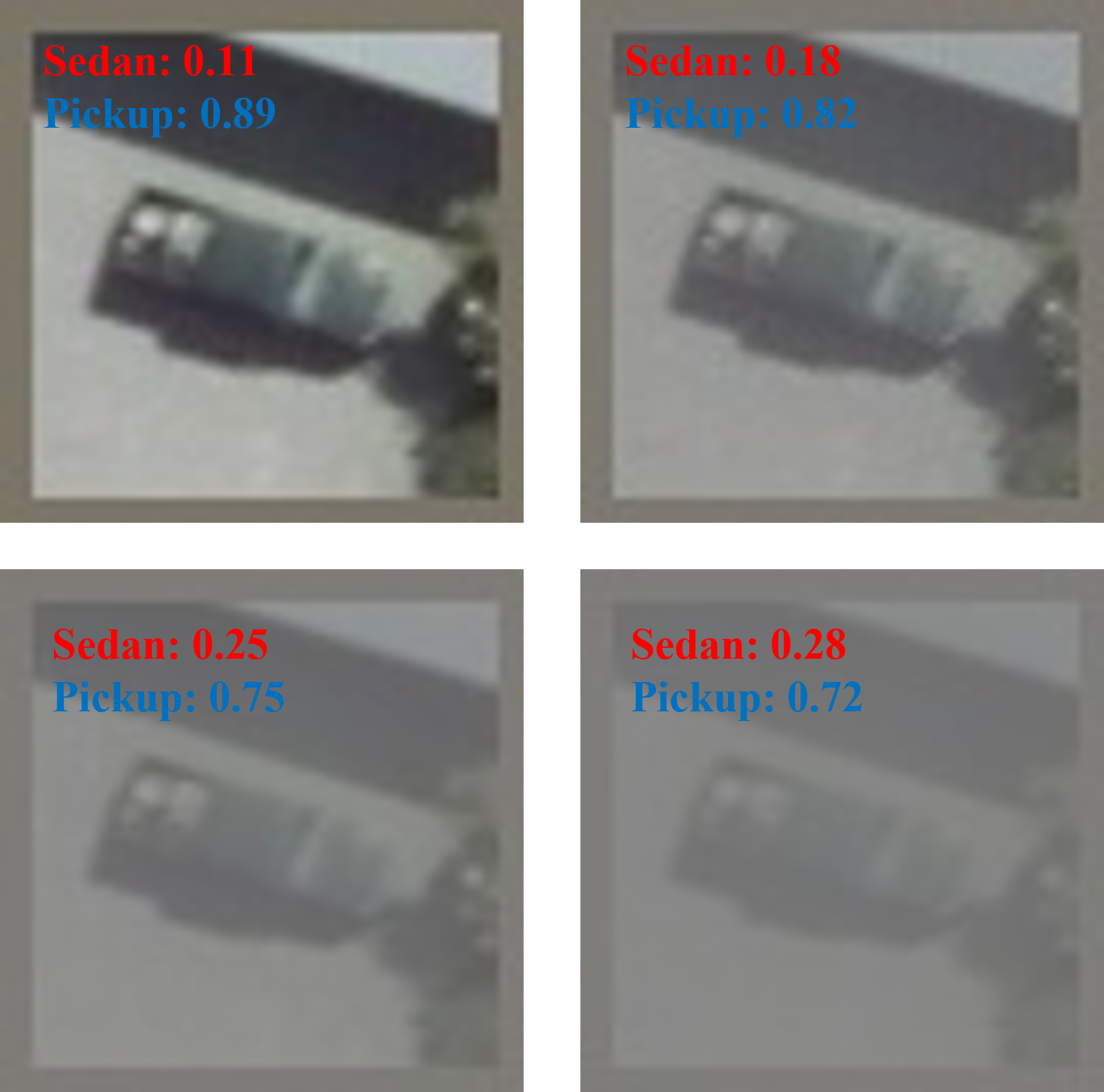}
	\caption{Classification results with corresponding probabilities under the foggy weather condition.}
	\label{fig32}
\end{figure}
In this experiment, we keep evaluation criteria the same as part A and use all the 6 models, Dense-TNT s24, Dense-TNT s12, PoolFormer s24, PoolFormer s12, ViT l12 and ViT l2, to evaluate the classification ability of Dense-TNT when the input data are affected by different weather conditions. Table \uppercase\expandafter{\romannumeral5} shows the results of the experiment. Fig. 7 shows the classification results with probabilities after processing under the foggy weather condition.

Fig. 8 also shows F1-scores of the experiment in histogram. Similarly, Dense-TNT, especially Dense-TNT s24 has better performance than other baselines. When the fog is heavier, the model leads other baselines by more.

\begin{figure}[!htb]
	\centering
	\includegraphics[width=1\linewidth]{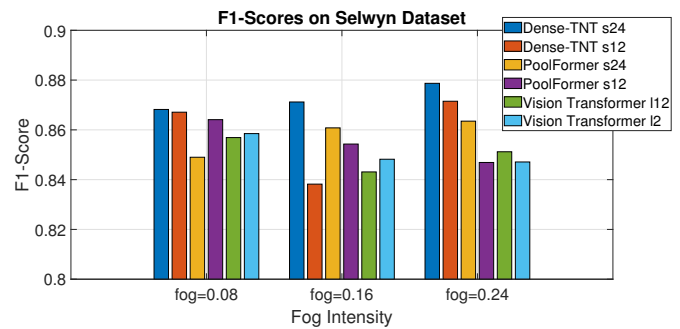}
	\caption{F1-scores of experimental results for Selwyn dataset under differebt foggy condition.}
	\label{fig34}
\end{figure}

With the data input affected by different levels of foggy weather, there is a certain level of decay in the accuracy of all the 6 models. Even though the decrease of accuracy, Dense-TNT s24 still has a relatively better performance than PoolFormer s24 and ViT l12. Dense-TNT s12 also has a generally better performance than PoolFormer s12 and ViT l2, which means that Dense-TNT can still be useful when dealing with affected input data.

\section{Conclusion}
In this paper, a novel classification neural network called Dense-TNT is firstly proposed to recognize vehicle type based on satellite remote sensing imagery. Dense-TNT is combined with DenseNet layer and TNT layer to capture both locality and global information from input images and is expected to achieve better recognition performance than other widely used methods, especially under complex weather conditions. Related experiments are designed to validate the feasibility of Dense-TNT based on the real-world remote sensing dataset collected from three different regions over multi-environment states. Necessary data preprocessing on the same dataset is executed to imitate the foggy weather condition in three different degrees: light, medium and heavy foggy. The experiment results show that Dense-TNT achieves better recognition performance with around 5$\%$-10$\%$ accuracy improvement than baseline algorithms, including PoolFormer and ViT. Under foggy weather conditions, the improvement is even larger. To summarize, the vehicle type classification performance of the proposed Dense-TNT framework under comprehensive weather conditions using remote sensing imagery has been verified.

\section*{Acknowledgment}
This study is supported by the RIE2020 Industry Alignment Fund – Industry Collaboration Projects (IAF-ICP) Funding Initiative, as well as cash and in-kind contribution from the industry partner(s) and A*STAR by its RIE2020 Advanced Manufacturing and Engineering (AME) Industry Alignment Fund C Pre-Positioning (IAF-PP) (Award A19D6a0053).

\ifCLASSOPTIONcaptionsoff
  \newpage
\fi


\singlespacing
\bibliographystyle{ieeetran}
\bibliography{autosam}


\begin{IEEEbiography}[{\includegraphics[width=1in,height=1.25in,clip,keepaspectratio]{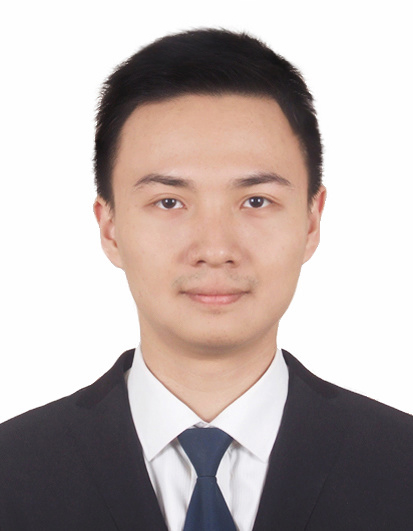}}]{Ruikang Luo}
    received the B.E. degree from the School of Electrical and Electronic Engineering, Nanyang Technological University, Singapore. He is currently currently pursuing the Ph.D. degree with the School of Electrical and Electronic Engineering, Nanyang Technological University, Singapore. His research interests include long-term traffic forecasting based on spatiotemporal data and artificial intelligence.
\end{IEEEbiography}

\begin{IEEEbiography}[{\includegraphics[width=1in,height=1.25in,clip,keepaspectratio]{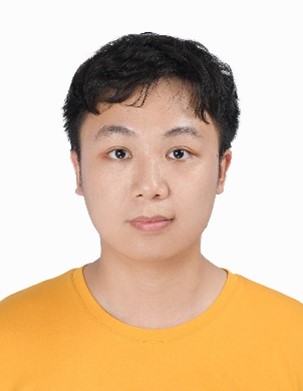}}]{Yaofeng Song}
	received the bachelor degree from the school of Automation Science and Engineering in South China University of Technology. Currently he is a Msc student in the school of Electrical and Electronic Engineering in Nanyang Technological University, Singapore. His research interests invlove deep learning based traffic forecasting.
\end{IEEEbiography}

\begin{IEEEbiography}[{\includegraphics[width=1in,height=1.25in,clip,keepaspectratio]{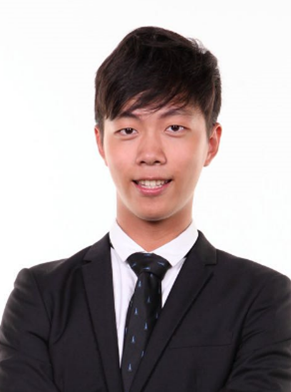}}]{Han Zhao}
	received the Bachelor degree in Electrical and Electronic Engineering from Nanyang Technological University, Singapore in 2018. He is currently working toward the Ph.D degree in Electrical and Electronic Engineering in Nanyang Technological University, Singapore. His research interests include intelligent transportation system (ITS), short-term traffic flow prediction and graph neural networks.
\end{IEEEbiography}

\begin{IEEEbiography}[{\includegraphics[width=1in,height=1.25in,clip,keepaspectratio]{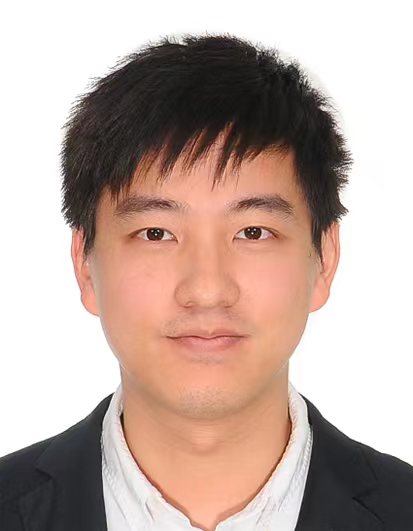}}]{Yicheng Zhang}
	Yicheng Zhang received the Bachelor of Engineering in Automation from Hefei University of Technology in 2011, the Master of Engineering degree in Pattern Recognition and Intelligent Systems from University of Science and Technology of China in 2014, and the PhD degree in Electrical and Electronic Engineering from Nanyang Technological University, Singapore in 2019. He is currently a research scientist at the Institute for Infocomm Research (I2R) in the Agency for Science, Technology and Research, Singapore (A*STAR). Before joining I2R, he was a research associate affiliated with Rolls-Royce @ NTU Corp Lab. He has participated in many industrial and research projects funded by National Research Foundation Singapore, A*STAR, Land Transport Authority, and Civil Aviation Authority of Singapore. He published more than 70 research papers in journals and peer-reviewed conferences. He received the IEEE Intelligent Transportation Systems Society (ITSS) Young Professionals Traveling Scholarship in 2019 during IEEE ITSC, and as a team member, received Singapore Public Sector Transformation Award in 2020.
\end{IEEEbiography}

\begin{IEEEbiography}[{\includegraphics[width=1in,height=1.25in,clip,keepaspectratio]{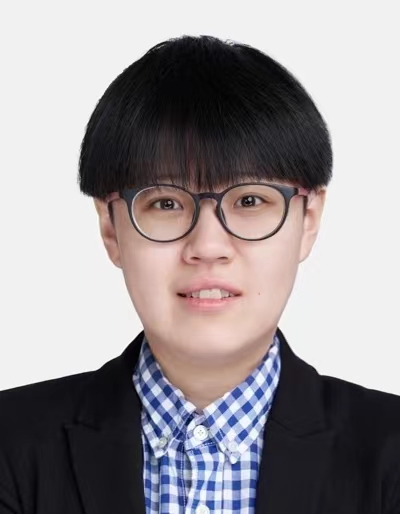}}]{Yi Zhang}
	received her Bachelor degree of Engineering from Shandong University, China in 2014, and the PhD degree in Electrical and Electronic Engineering from Nanyang Technological University, Singapore in 2020. She is currently a research scientist at the Institute for Infocomm Research (I2R) in the Agency for Science, Technology and Research, Singapore (A*STAR). Her research interests focus on intelligent transportation system, including urban traffic flow management, model-based traffic signal scheduling, lane change prediction and bus dispatching and operation management.
\end{IEEEbiography}

\begin{IEEEbiography}[{\includegraphics[width=1in,height=1.25in,clip,keepaspectratio]{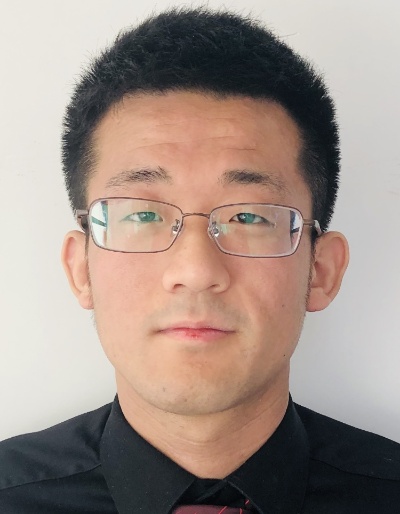}}]{Nanbin Zhao}
	received the Bachelor’s Degree of Engineering from University of Electronic Science and Technology of China in 2019 and the Master’s Degree of Science from the National University of Singapore in 2020. He is currently pursuing the Ph.D. degree at the School of Electrical and Electronic Engineering, Nanyang Technological University. 
	His research interests include intelligent transportation systems, vehicle control, machine learning, and IOT.
\end{IEEEbiography}

\begin{IEEEbiography}[{\includegraphics[width=1in,height=1.25in,clip,keepaspectratio]{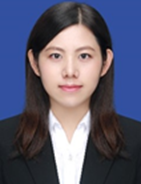}}]{Liping Huang}
	obtained her Ph. D, and Master of Computer Science from Jilin University in 2018 and 2014, respectively. She has been working as a research fellow at Nanyang Technological University since 2019 June. Dr. Huang’s research interests include spatial and temporal data mining, mobility data pattern recognition, time series prediction, machine learning, and job shop scheduling. In the aforementioned areas, she has more than twenty publications and serves as the reviewer of multiple journals, such as IEEE T-Big Data, IEEE T-ETCI, et al.
\end{IEEEbiography}

\begin{IEEEbiography}[{\includegraphics[width=1in,height=1.25in,clip,keepaspectratio]{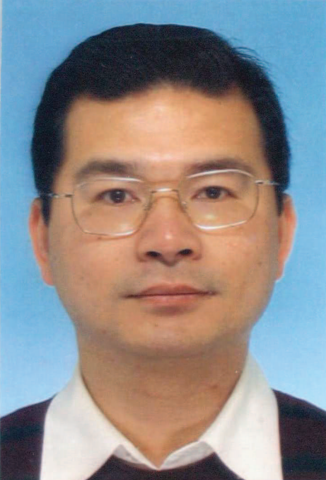}}]{Rong Su}
	received the M.A.Sc. and Ph.D. degrees both	in electrical engineering from the University of Toronto, Toronto, Canada, in 2000 and 2004 respectively.	He is affiliated with the School of Electrical and Electronic Engineering, Nanyang Technological University, Singapore. His research interests include modeling, fault diagnosis and supervisory control of discrete-event dynamic systems. Dr. Su has been a member of IFAC technical committee on discrete event and hybrid systems (TC 1.3) since 2005.
\end{IEEEbiography}

\end{document}